\documentclass[10pt,twocolumn,letterpaper]{article}

\usepackage{wacv}
\usepackage{times}
\usepackage{epsfig}
\usepackage{graphicx}
\usepackage{amsmath}
\usepackage{amssymb}
\usepackage[accsupp]{axessibility}  % Improves PDF readability for those with disabilities.

\usepackage{multirow}
\usepackage{booktabs}
\usepackage{xcolor}
\usepackage{makecell}
\usepackage[ruled,vlined]{algorithm2e}
\usepackage[numbers,sort,compress]{natbib}

\def\wacvPaperID{637} % Enter the WACV Paper ID here

\wacvfinalcopy % *** Uncomment this line for the final submission

\ifwacvfinal
\fi

\ifwacvfinal
\usepackage[breaklinks=true,bookmarks=false]{hyperref}
\else
\usepackage[pagebackref=true,breaklinks=true,colorlinks,bookmarks=false]{hyperref}
\fi

\newcommand\blfootnote[1]{%
  \begingroup
  \renewcommand\thefootnote{}\footnote{#1}%
  \addtocounter{footnote}{-1}%
  \endgroup
}
\begin{document}

%%%%%%%%% TITLE
\title{Hierarchical Proxy-based Loss for Deep Metric Learning}

\author{Zhibo Yang\textsuperscript{1*}, Muhammet Bastan\textsuperscript{2}, Xinliang Zhu\textsuperscript{2}, Doug Gray\textsuperscript{2}, Dimitris Samaras\textsuperscript{1} \\
\textsuperscript{1}Stony Brook University,~~~  \textsuperscript{2}Visual Search \& AR, Amazon
}

\maketitle

\ifwacvfinal
\thispagestyle{empty}
\fi
\def\mA{\mathcal{A}}
\def\mB{\mathcal{B}}
\def\mC{\mathcal{C}}
\def\mD{\mathcal{D}}
\def\mE{\mathcal{E}}
\def\mF{\mathcal{F}}
\def\mG{\mathcal{G}}
\def\mH{\mathcal{H}}
\def\mI{\mathcal{I}}
\def\mJ{\mathcal{J}}
\def\mK{\mathcal{K}}
\def\mL{\mathcal{L}}
\def\mM{\mathcal{M}}
\def\mN{\mathcal{N}}
\def\mO{\mathcal{O}}
\def\mP{\mathcal{P}}
\def\mQ{\mathcal{Q}}
\def\mR{\mathcal{R}}
\def\mS{\mathcal{S}}
\def\mT{\mathcal{T}}
\def\mU{\mathcal{U}}
\def\mV{\mathcal{V}}
\def\mW{\mathcal{W}}
\def\mX{\mathcal{X}}
\def\mY{\mathcal{Y}}
\def\mZ{\mathcal{Z}} 

\def\bbN{\mathbb{N}} 
\def\bbR{\mathbb{R}} 
\def\bbP{\mathbb{P}} 
\def\bbQ{\mathbb{Q}} 
\def\bbE{\mathbb{E}}

\def\1n{\mathbf{1}_n}
\def\0{\mathbf{0}}
\def\1{\mathbf{1}}

\def\A{{\bf A}}
\def\B{{\bf B}}
\def\C{{\bf C}}
\def\D{{\bf D}}
\def\E{{\bf E}}
\def\F{{\bf F}}
\def\G{{\bf G}}
\def\H{{\bf H}}
\def\I{{\bf I}}
\def\J{{\bf J}}
\def\K{{\bf K}}
\def\L{{\bf L}}
\def\M{{\bf M}}
\def\N{{\bf N}}
\def\O{{\bf O}}
\def\P{{\bf P}}
\def\Q{{\bf Q}}
\def\R{{\bf R}}
\def\S{{\bf S}}
\def\T{{\bf T}}
\def\U{{\bf U}}
\def\V{{\bf V}}
\def\W{{\bf W}}
\def\X{{\bf X}}
\def\Y{{\bf Y}}
\def\Z{{\bf Z}}

\def\a{{\bf a}}
\def\b{{\bf b}}
\def\c{{\bf c}}
\def\d{{\bf d}}
\def\e{{\bf e}}
\def\f{{\bf f}}
\def\g{{\bf g}}
\def\h{{\bf h}}
\def\i{{\bf i}}
\def\j{{\bf j}}
\def\k{{\bf k}}
\def\l{{\bf l}}
\def\m{{\bf m}}
\def\n{{\bf n}}
\def\o{{\bf o}}
\def\p{{\bf p}}
\def\q{{\bf q}}
\def\r{{\bf r}}
\def\s{{\bf s}}
\def\t{{\bf t}}
\def\u{{\bf u}}
\def\v{{\bf v}}
\def\w{{\bf w}}
\def\x{{\bf x}}
\def\y{{\bf y}}
\def\z{{\bf z}}

\def\balpha{\mbox{\boldmath{$\alpha$}}}
\def\bbeta{\mbox{\boldmath{$\beta$}}}
\def\bdelta{\mbox{\boldmath{$\delta$}}}
\def\bgamma{\mbox{\boldmath{$\gamma$}}}
\def\blambda{\mbox{\boldmath{$\lambda$}}}
\def\bsigma{\mbox{\boldmath{$\sigma$}}}
\def\btheta{\mbox{\boldmath{$\theta$}}}
\def\bomega{\mbox{\boldmath{$\omega$}}}
\def\bxi{\mbox{\boldmath{$\xi$}}}
\def\bnu{\mbox{\boldmath{$\nu$}}}                                  
\def\bphi{\mbox{\boldmath{$\phi$}}}
\def\bmu{\mbox{\boldmath{$\mu$}}}

\def\bDelta{\mbox{\boldmath{$\Delta$}}}
\def\bOmega{\mbox{\boldmath{$\Omega$}}}
\def\bPhi{\mbox{\boldmath{$\Phi$}}}
\def\bLambda{\mbox{\boldmath{$\Lambda$}}}
\def\bSigma{\mbox{\boldmath{$\Sigma$}}}
\def\bGamma{\mbox{\boldmath{$\Gamma$}}}
                                  
\newcommand{\myprob}[1]{\mathop{\mathbb{P}}_{#1}}

\newcommand{\myexp}[1]{\mathop{\mathbb{E}}_{#1}}

\newcommand{\mydelta}[1]{1_{#1}}

\newcommand{\myminimum}[1]{\mathop{\textrm{minimum}}_{#1}}
\newcommand{\mymaximum}[1]{\mathop{\textrm{maximum}}_{#1}}    
\newcommand{\mymin}[1]{\mathop{\textrm{minimize}}_{#1}}
\newcommand{\mymax}[1]{\mathop{\textrm{maximize}}_{#1}}
\newcommand{\mymins}[1]{\mathop{\textrm{min.}}_{#1}}
\newcommand{\mymaxs}[1]{\mathop{\textrm{max.}}_{#1}}  
\newcommand{\myargmin}[1]{\mathop{\textrm{argmin}}_{#1}} 
\newcommand{\myargmax}[1]{\mathop{\textrm{argmax}}_{#1}} 
\newcommand{\myst}{\textrm{s.t. }}

\newcommand{\denselist}{\itemsep -1pt}
\newcommand{\sparselist}{\itemsep 1pt}

\definecolor{pink}{rgb}{0.9,0.5,0.5}
\definecolor{purple}{rgb}{0.5, 0.4, 0.8}   
\definecolor{gray}{rgb}{0.3, 0.3, 0.3}
\definecolor{mygreen}{rgb}{0.2, 0.6, 0.2}

\newcommand{\cyan}[1]{\textcolor{cyan}{#1}}
\newcommand{\red}[1]{\textcolor{red}{#1}}  
\newcommand{\blue}[1]{\textcolor{blue}{#1}}
\newcommand{\magenta}[1]{\textcolor{magenta}{#1}}
\newcommand{\pink}[1]{\textcolor{pink}{#1}}
\newcommand{\green}[1]{\textcolor{green}{#1}} 
\newcommand{\gray}[1]{\textcolor{gray}{#1}}    
\newcommand{\mygreen}[1]{\textcolor{mygreen}{#1}}    
\newcommand{\purple}[1]{\textcolor{purple}{#1}}       

\definecolor{greena}{rgb}{0.4, 0.5, 0.1}
\newcommand{\greena}[1]{\textcolor{greena}{#1}}

\definecolor{bluea}{rgb}{0, 0.4, 0.6}
\newcommand{\bluea}[1]{\textcolor{bluea}{#1}}
\definecolor{reda}{rgb}{0.6, 0.2, 0.1}
\newcommand{\reda}[1]{\textcolor{reda}{#1}}

\def\changemargin#1#2{\list{}{\rightmargin#2\leftmargin#1}\item[]}
\let\endchangemargin=\endlist
                                               
\newcommand{\cm}[1]{}

\newcommand{\mhoai}[1]{{\color{purple}\textbf{[MH: #1]}}}

\newcommand{\mtodo}[1]{{\color{red}$\blacksquare$\textbf{[TODO: #1]}}}
\newcommand{\myheading}[1]{\vspace{1ex}\noindent \textbf{#1}}
\newcommand{\htimesw}[2]{\mbox{$#1$$\times$$#2$}}

% The following are useful for creating homework or exams

\newif\ifshowsolution
%\showsolutionfalse
\showsolutiontrue

\ifshowsolution  
\newcommand{\Comment}[1]{\paragraph{\bf $\bigstar $ COMMENT:} {\sf #1} \bigskip}
\newcommand{\Solution}[2]{\paragraph{\bf $\bigstar $ SOLUTION:} {\sf #2} }
\newcommand{\Mistake}[2]{\paragraph{\bf $\blacksquare$ COMMON MISTAKE #1:} {\sf #2} \bigskip}
\else
\newcommand{\Solution}[2]{\vspace{#1}}
\fi

\newcommand{\truefalse}{
\begin{enumerate}
	\item True
	\item False
\end{enumerate}
}

\newcommand{\yesno}{
\begin{enumerate}
	\item Yes
	\item No
\end{enumerate}
}

\newcommand{\Sref}[1]{Sec.~\ref{#1}}
\newcommand{\Eref}[1]{Eq.~(\ref{#1})}
\newcommand{\Fref}[1]{Fig.~\ref{#1}}
\newcommand{\Tref}[1]{Table~\ref{#1}}

\begin{abstract}
  % motivation problem approach results conclusion (impact)
%   In deep metric learning (DML), we are interested in learning to embed images as feature vectors whose distances are useful for tasks such as image retrieval. A number of loss functions have been proposed for Deep Metric Learning (DML) in recent years, among which, proxy-based losses are superior in their fast convergence and low training complexity.
  Proxy-based metric learning losses are superior to pair-based losses due to their fast convergence and low training complexity. However, existing proxy-based losses focus on learning class-discriminative features while overlooking the commonalities shared across classes which are potentially useful in describing and matching samples. Moreover, they ignore the implicit hierarchy of categories in real-world datasets, where similar subordinate classes can be grouped together. In this paper, we present a framework that leverages this implicit hierarchy by imposing a hierarchical structure on the proxies and can be used with any existing proxy-based loss. This allows our model to capture both class-discriminative features and class-shared characteristics without breaking the implicit data hierarchy. We evaluate our method on five established image retrieval datasets such as In-Shop and SOP. Results demonstrate that our hierarchical proxy-based loss framework improves the performance of existing proxy-based losses, especially on large datasets which exhibit strong hierarchical structure.
  \blfootnote{*Work partially done while interning at Amazon.}
\end{abstract}

\section{Introduction}
% overview of DML
Learning visual similarity has many important applications in computer vision, ranging from image retrieval \cite{ge2020self} to video surveillance (e.g., person re-identification \cite{dai2019batch}). It is often treated as a metric learning problem where the task is to represent images with compact embedding vectors such that semantically similar images are grouped together while dissimilar images are far apart in the embedding space. Inspired by the success of deep neural networks in visual recognition, convolutional neural networks have also been employed in metric learning, which is specifically called deep metric learning (DML) \cite{boudiaf2020unifying, chen2017beyond, dai2019batch, elezi2020group, levi2020reducing, oh2016deep, musgrave2020metric,wu2017sampling}.

\begin{figure}[t]
  \centering
  \includegraphics[width=1.0\linewidth]{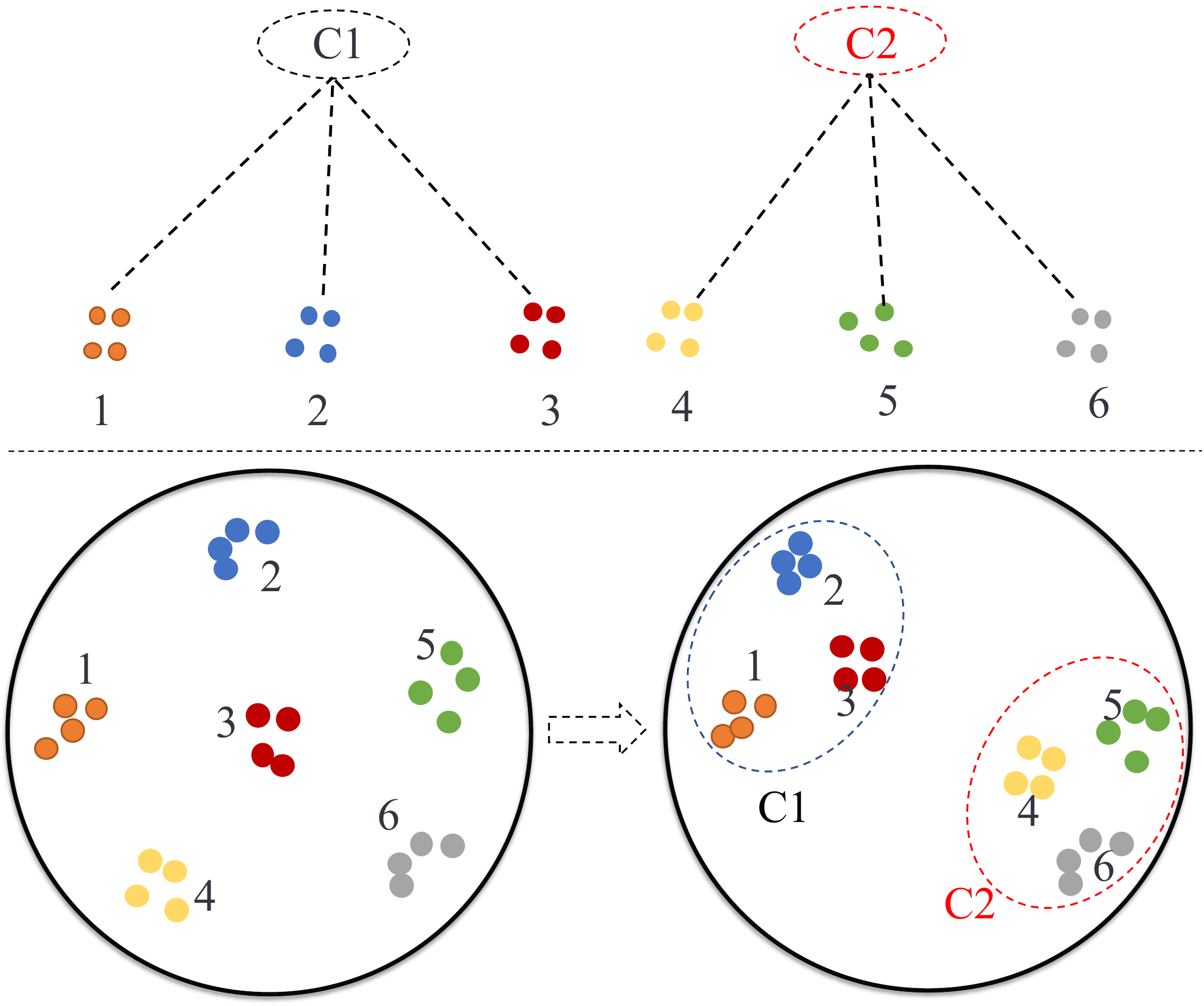}
  \caption{\textbf{Hierarchical proxies}. Traditional proxy-based losses seek to learn an embedding space where each class is well separated (see bottom-left where different colors denote different classes). However, real-world data often have implicit hierarchies. For example, classes 1-3 might belong to one super category while classes 4-6 belong to another (top panel). Our proposed HPL learns to separate the classes and captures this hierarchy to pull samples in the same super category closer (bottom-right).}
  \label{fig:hps}
\end{figure}

In recent years, a number of DML loss functions \cite{kim2020proxy,schroff2015facenet,movshovitz2017no} have been developed to guide network training for visual similarity learning. 
%These loss functions can be broadly categorized into two groups: pair-based and proxy-based. 
The two dominant groups of these DML loss functions are pair-based losses and proxy-based losses.
Pair-based losses (e.g., contrastive loss \cite{hadsell2006dimensionality} and triplet loss \cite{schroff2015facenet}) directly compute the loss based on pairs of samples with the goal of encouraging samples in the same class to be close to each other and samples from different classes to be far apart. This is different from classification losses which are computed individually for each training sample. Pair-based losses compute loss for each tuple of samples formed in a training batch. Hence, a major drawback of pair-based losses is that for a fixed number $N$ of training samples there is a prohibitively large number of tuples (i.e., $O(N^2)$ or $O(N^3)$) including many non-informative tuples, which leads to slow convergence and degraded performance.
In contrast, proxy-based losses (e.g., proxy-NCA \cite{movshovitz2017no} and proxy anchor loss \cite{kim2020proxy}) try to learn a set of data points, called \textit{proxies}, to approximate the data space of the training set. 
At each iteration, triplets are formed between samples from a local training batch and the global proxies to train the embedding networks as well as to update the proxies.
Since the number of proxies is often orders of magnitude smaller than the number of samples in the training set, proxy-based losses significantly reduce the high training complexity of pair-based losses.

In the image retrieval context, where the task is to find visually similar images in a large gallery of seen and unseen classes given a query image, real-world datasets, like SOP \cite{oh2016deep}, ImageNet \cite{ILSVRC15}, often contain an implicit hierarchy of categories, e.g., huskies can be categorized as spitzs, as simply dogs, or more broadly as animals. Common features shared across the finer-grained classes often characterize their superordinate category. However, existing proxy-based losses ignore this hierarchy, focusing on extracting class-discriminative features and overlooking features that are shared across classes that could be useful for describing and retrieving images. Taking dogs as an example, collies and German shepherds are medium-sized dogs with thin snouts and upright ears, while collies are distinctive from German shepherds in their long hair. Given a training set of collie and German shepherd images, a typical proxy-based model might only learn features related to long hair while overlooking features shared among classes such as medium size which is helpful in discerning collies from shelties.

In this paper, we propose a simple method that imposes a hierarchical structure on the proxies (see \Fref{fig:hps} for illustration) and can be combined with existing proxy-based losses. We name our method \textit{Hierarchical Proxy-based Loss (HPL)}. Building on top of existing proxy-based losses, our HPL creates a pyramid of learnable proxies where the lowest/finest level of proxies are created similarly to the existing proxy-based losses (i.e, one proxy per class). The higher-level proxies are learned in an unsupervised manner. Each coarser level of proxies are the cluster centroids of its lower level of proxies. Hence, each sample is associated with one proxy at every level in the proxy pyramid and the losses are computed for each level independently. 
% There are two major merits of HPL: 1) the coarse proxies serve as super categories which breaks the boundary of classes and thus help to learn features shared across classes; 2) real-world data/categories are often organized in hierarchy (e.g., ImageNet \cite{ILSVRC15} is organized according to the WordNet \cite{miller1998wordnet} hierarchy),
%or a product catalog where product images are organized in a hierarchy of catalog categories.
% and the inbuilt proxy hierarchy in our HPL can potentially better characterize the data space, hence less prone to overfitting.
In experiments, we demonstrate that our HPL improves traditional proxy-based loss performance on several public image retrieval benchmarks including Stanford Online Products (SOP) and In-Shop Clothes Retrieval (In-Shop), CUB200, Cars-196 and iNaturalist. Verifying our initial motivation, larger performance improvements are observed for datasets with a larger number of classes and a more complex hierarchical structure---Recall@1 is increased by +2.50\% on SOP and +2.87\% on In-Shop. Performance still matches the state-of-the-art in the case of the CUB200 dataset which contains 200 fine-grained bird categories with subtle differences, lacking a strong hierarchy.
%as shown in figure \ref{fig:vis_sop}. 

% contribution to be finalized
In summary, our  contributions  are: 1) We propose a novel hierarchical proxy-based method which is applicable to all existing proxy-based losses; 2) We demonstrate that the proposed method improves the performance of traditional proxy-based losses on several image retrieval benchmarks, especially for datasets with strong and complex hierarchical structure; 3) We also show that the hierarchy learned by our method outperforms a human-curated hierarchy for image retrieval.

\section{Related Work}

\myheading{Proxy-based losses.}
% Similar to classification networks which connect the whole dataset using the final weight matrix that maps embeddings to a universal vector of class logits, proxy-based losses create a set of learnable proxies shared across all samples and model the image-to-proxy relations instead of image-to-image relations in pair-based losses. Hence, the loss can be computed independently for each sample in a batch as in the classification losses. 
Proxy-NCA \cite{movshovitz2017no}, the first proxy-based DML loss,  addressed the slow convergence of pair-based losses (e.g. Triplet loss \cite{schroff2015facenet}). The main idea is to create one proxy for each class (by default) and use the Neighborhood Component Analysis (NCA) loss \cite{roweis2004neighbourhood} to pull samples close to its assigned proxy while pushing samples away from non-assigned proxies. Proxy-NCA++ \cite{teh2020proxynca++} further improves Proxy-NCA with several training improvements. Proxy Anchor \cite{kim2020proxy} uses proxies as anchors to leverage the rich image-to-image relations that are missing in Proxy-NCA. Other methods like SoftTriple loss \cite{qian2019softtriple} and Manifold Proxy loss \cite{aziere2019ensemble} respectively extend the Softmax  and N-pair losses to  proxy-based versions. To an extent, all these proxy-based losses treat deep metric learning as a classification task by using the proxies to separate samples from different classes, and thus they focus on extracting class-discriminative features. In contrast, our proposed HPL builds on existing proxy-based losses and explicitly models the class-shared features by imposing a hierarchical structure in the proxies. This helps to regularize the embeddings and prevent the model from overfitting to the training classes.

\myheading{Modeling the data distribution.}
Similar to our approach, hierarchical triplet loss (HTL) \cite{ge2018deep} and HiBsteR \cite{waltner2019hibster} also try to leverage the underlying data hierarchy for DML. However, HTL has been developed for the Triplet loss and uses the data hierarchy to mine good training samples, while our method is designed for proxy-based losses and aims to learn class-shared information by modeling the data hierarchy.
HiBsteR requires ground-truth hierarchical labels which limits its applicability, while our method does not.
Divide and Conquer (DC) \cite{sanakoyeu2019divide} tries to adapt the embeddings to a nonuniform data distribution, while we aim to learn embeddings that capture the underlying data hierarchy. Moreover, DC needs to cluster the entire dataset repeatedly which can be prohibitively expensive for large datasets, while our method operates on proxies which are often orders of magnitude smaller than the whole dataset.
% Moreover, both DC and HiBsteR learn an ensemble of equal-sized small embeddings, aiming to better approximate the data landscape but by fragmenting the representation they may fail to capture the discriminative features given the limited capacity of the shorter-length representations of the embeddings. In contrast, our method learns a single metric which not only adapts well to the data hierarchy but also maintains a good separation among classes. 
In addition, our HPL also resembles PIEs \cite{ho2019pies} in learning from different groups of data, but PIEs aims to learn a pose-invariant embedding from different views of objects in each class and requires additional pose labels while our HPL learns from groups of classes and does not require additional annotation.

\myheading{Modeling class-shared information.}
MIC \cite{roth2019mic} and DiVA \cite{milbich2020diva} also try to learn features shared across classes. However, both MIC and DiVA formulate a multi-task learning problem and seek to learn separate embeddings for class-specific information and class-shared information. Instead, our HPL imposes additional regularization on top of the original proxy-based losses by explicitly modeling class-shared information. Moreover, MIC clusters over the entire dataset which is not scalable to large datasets, while we cluster on the proxies whose number is much smaller than the size of the entire dataset. DiVA is based on triplet loss while our HPL is developed for proxy-based losses and is compatible with multiple proxy-based losses.
Our HPL also shares a similar motivation with another line of research \cite{yan2015hd, kim2018hierarchy, murthy2016deep} which aims to design better image classifiers by leveraging the hierarchical structure in data. However, they focus on the network architectures design to better separate the classes, while our work aims to design a better metric learning loss function which takes advantages of class-shared information. 

\section{Method}

In this section, we  briefly review two popular proxy-based losses: Proxy-NCA and Proxy Anchor. Then we introduce our hierarchical proxy-based loss.
\subsection{Preliminaries}
Given a training set $\mathcal{D}=\{(x_i, y_i)\}_{i=1}^N$ where $N$ is the number of data samples, $x_i$ and $y_i$ are the $i$-th training image and its class label, respectively. The goal of deep metric learning is to learn a similarity function $s(x_i, x_j)$ such that 
\begin{equation}
    s(x_i, x_j) \ge s(x_i, x_k)\quad \forall i, j, k, y_i=y_j\neq y_k.
    \label{eq:dml}
\end{equation}
In DML, the similarity function is often learned via a deep neural network parameterized by $\theta$ which maps the image $x_i$ to an embedding vector $\Phi_\theta(x_i)$. For simplicity, we omit $\Phi$ in the notation and denote the similarity function with $s(x_i, x_j;\theta)$ which is equivalent to $s(\Phi_\theta(x_i),\Phi_\theta(x_j))$. In the following, $x_i$ and $\Phi_\theta(x_i)$ will be used interchangeably. There are two popular similarity function choices which are the Euclidean distance, i.e., $s(x_i, x_j)=-||x_i-x_j||_2$ and the cosine similarity, i.e., $s(x_i, x_j)=\frac{x_i\cdot x_j}{||x_i||_2||x_j||_2}$.

\subsection{Existing Proxy-based Losses}
Similar to a classification network whose last weight matrix connects all samples in the training set by mapping them to a universal vector of class logits, proxy-based losses link samples across batches by using a set of learnable proxies $P=\{p_i\}_{i=1}^C$. One proxy is usually created for each class\footnote{The analysis can also carry over for losses that assign multiple proxies to a class such as SoftTriple \cite{qian2019softtriple}.}, hence, $C$ is the number of classes in the training set. Each proxy is a vector of the same size as the embedding vector. Hence, \Eref{eq:dml} is transformed as 
\begin{equation}
    s(x_i,p_{y_j};\theta,P)\ge s(x_i,p_{y_k};\theta,P), \forall i, j, k, y_i=y_j\neq y_k.
    \label{eq:proxy}
\end{equation}

Proxy-NCA \cite{movshovitz2017no} optimizes $P$ and $\theta$ by minimizing the following approximated NCA \cite{roweis2004neighbourhood} loss
\begin{equation}
    \ell(X, Y, P)=-\sum_{(x_i, y_i)\in B}\log \frac{e^{s(x_i, p_{y_i})}}{\sum_{y_j\neq y_i}e^{s(x_i, p_{y_j})}},
    \label{eq:pnca}
\end{equation}
where $B=\{X, Y\}$ is a batch of training samples and $s(\cdot,\cdot)$ is the cosine similarity function.

As opposed to Proxy-NCA which uses images as anchors, Proxy Anchor \cite{kim2020proxy} uses proxies as anchors and forms a positive set $X_p^+$ and a negative set $X_p^-$ from samples in a batch for each proxy $p$ according to their class labels:
\begin{equation}
\begin{array}{cc}
     \ell(X, Y, P) = & \frac{1}{|P_B|}\sum\limits_{p\in P_B}\log \big(1+\sum\limits_{x\in X_p^+} e^{-\alpha (s(x,p)-\delta)}\big) \\
     & + \frac{1}{|P|}\sum\limits_{p\in P}\log\big(1+\sum\limits_{x\in X_p^-} e^{\alpha (s(x,p)+\delta)}\big),
     \label{eq:pa}
\end{array}
\end{equation}
where $\alpha$ and $\delta$ are the scaling factor and the margin, respectively. $P_B$ denotes the set of proxies with at least one positive sample in the batch $B$.

Different from classification tasks, it is crucial for the learned embeddings to generalize to unseen classes in DML. Therefore, although the use of proxies reduces the high sampling complexity, it raises another issue: class-specific proxies make the embedding networks focus only on learning class-discriminative features while overlooking class-shared information, which makes the learned representation vulnerable to overfitting and undermines its generalization to unseen classes. To overcome this shortcoming, we introduce the hierarchical proxy-based loss below.

\begin{figure*}[t]
  \centering
  \includegraphics[width=.9\linewidth]{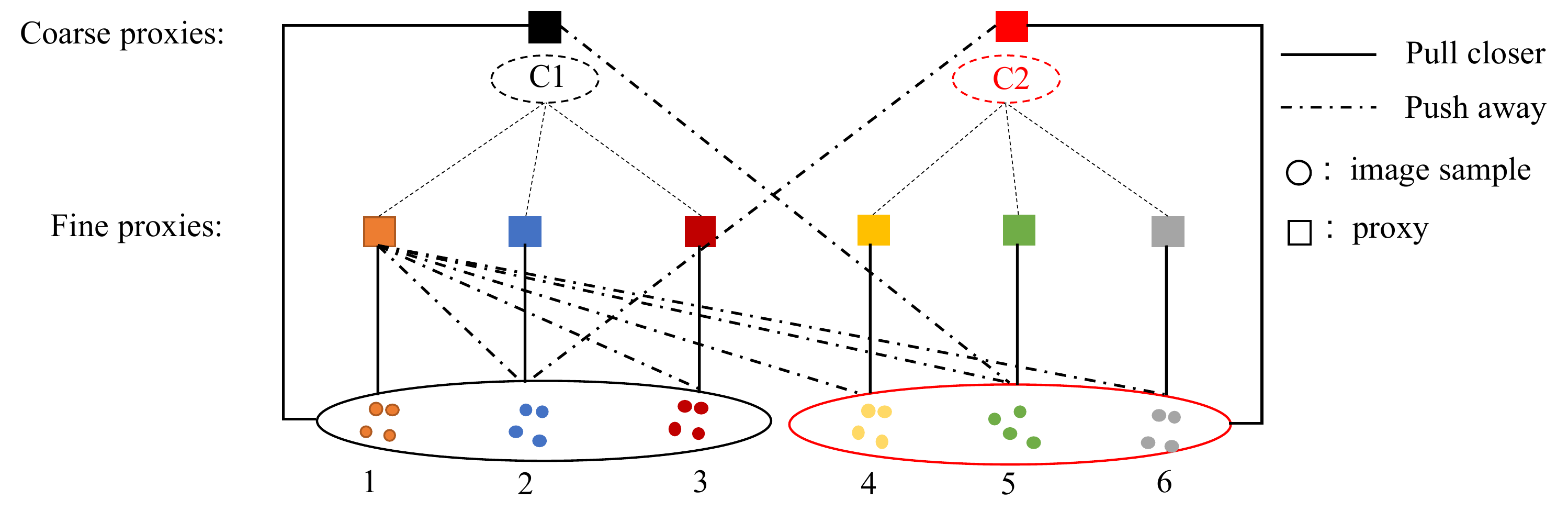}
  \caption{\textbf{The framework of our hierarchical proxy-based loss}. HPL builds on top of existing proxy-based losses and computes the loss for each level of proxies separately. HPL learns an embedding space with small within-class distance but also small within-cluster (i.e., coarse proxies) distance. For illustration clarity, the full interactions between fine proxies and samples are presented for only class 1, class 2-6 are similar. Each color represent a category or super category.}
  \label{fig:hpl}
  \vspace{-0.2cm}
\end{figure*}

\subsection{Hierarchical Proxy-based Loss}\label{sec:hpl}
To leverage the implicit data hierarchy, we enforce a hierarchical structure in the proxies by creating a proxy pyramid with $L$ levels of proxies $\mathcal{P}=\{P_0,\cdots,P_{L-1}\}$ where $(L\ge 1)$. The 0-th level of (fine) proxies $P_0$ are created the same as existing proxy-based losses---one proxy per class. Every proxy $p_i^l\in P_l$ is exclusively assigned to one proxy $p_j^{l+1}\in P_{l+1}$, where $0\le l<L-1$, 

Given a batch of samples $B=\{(x_i, y_i)\}$, each $x_i$ is associated with one and only one proxy at every level. In other words, a label $y_i^l$ is assigned to each sample $x_i$ at every level of proxies $P_l$. The label is assigned according to proxy membership. Namely, if $y_i^l=j$ and $p_j^l$ is assigned to $p_k^{l+1}$, then $y_i^{l+1}=k$. Note that $y_i^0=y_i$. During training, the loss is not only computed for the fine proxies at 0-th level, but also for all coarse proxies at higher level(s) (see  \Fref{fig:hpl} for illustration). Formally, our HPL seeks to learn a similarity function that satisfies 
\begin{equation}
\begin{array}{cc}
    s(x_i,p_{y_j^l};\theta,\mathcal{P})\ge s(x_i,p_{y_k^l};\theta,\mathcal{P}), \\\forall i, j, k, l, y_i^l=y_j^l\neq y_k^l.
    \label{eq:hp}
\end{array}
\end{equation}
Note that \Eref{eq:hp} is a generalization of \Eref{eq:proxy} and is equivalent to \Eref{eq:proxy} when $L=1$. Thus, we extend existing proxy-based losses and formulate the HPL loss as:
\begin{equation}
    \mathcal{L}=\sum_{l=0}^{L-1}\omega_l \ell(X, Y^l, P_l),
    \label{eq:hpl}
\end{equation}
where $\omega_l$ is the loss weight for the $i$-th level of proxies and $Y^l=\{y_i^l\}$. $\ell(X, Y^l,P_l)$ can be any existing proxy-based loss function such as Proxy-NCA loss (i.e., \Eref{eq:pnca}) and Proxy Anchor loss (i.e, \Eref{eq:pa}).

In this way, our HPL not only learns the class-discriminative features through the fine proxies as existing proxy-based losses do, but also captures class-shared information through higher-level coarse proxies. Moreover, the proxy hierarchy helps better represent the real-world data space, thus our HPL is more resilient to overfitting in real-world applications.

% how to learn coarse proxies
\subsection{Learning Proxies via Online Clustering}\label{sec:oc}
% Algorithm
\begin{algorithm}[t]
\SetAlgoLined
%  Train $P_0$ with traditional proxy-based loss\\
 \For{$0\le l<L-1$}{
 Cluster $P_l$ and set $P_{l+1}$ to be the cluster centroids and denote $Q_l$ the cluster assignment\\
 }
 \For{$k=1,2,\cdots,K$}{
 Sample a batch $B=\{x_i, y_i\}$ from $D$\\
 Assign labels to $x_i$ according to $Q=\{Q_l\}_{l=0}^{L-1}$: $B=\{x_i, y_i^0,\cdots, y_i^{L-1}\}$\\
  Update $\theta$ and $P_0$ with \Eref{eq:hpl}\\
  \If{$k\mod T == T - 1$}{
    \For{$0\le l<L-1$}{
         \For{$i=1,2,\cdots,|P_l|$}{
         Update the assignment:
         $Q_l^i=\arg\min_j||P_l^i - P_{l+1}^j||_2^2$\\
         }
         \For{$j=1,2,\cdots,|P_{l+1}|$}{
         Update the proxies:
         $P_{l+1}^j=\sum\limits_{Q_l^i=j} P_l^i\Big/|\mathbf{1}_{Q_l=j}|$\\
         }
    }
  }
 }
 \caption{HPL with Online Clustering}
 \label{alg:oc}
\end{algorithm}

Higher-level proxies (i.e., coarse proxies) represent super categories of the training data. Obviously, these coarse proxies can be similarly learned as the fine proxies if additional supervisory signals such as the super category labels are available. In this paper, we develop a method that works without additional supervision. Particularly, we create pseudo super categories by grouping the lower level of proxies into clusters using an unsupervised clustering algorithm (e.g., k-means), and use the cluster centroids as higher-level proxies. We adopt an online clustering approach \cite{zhan2020online} to update higher-level proxies.

We outline the training procedure in Algorithm \ref{alg:oc}. Starting with the fine proxies $P_0$, we perform offline k-means clustering on every lower-level proxy $P_l$ where $0\le l < L-1$ and use the cluster centroids to initialize higher-level proxies (i.e., $P_{l+1}$). At each iteration, we sample a batch of training data and compute the proxy labels $y_i^0, \cdots, y_i^{L-1}$ for each sample $x_i$ according to the mechanism described in \Sref{sec:hpl}. Then the weights of the embedding network $\theta$ and fine proxies $P_0$ are updated with our hierarchical proxy loss (i.e., \Eref{eq:hpl}) through back-propagation \cite{rumelhart1985learning}. Every $T$ iterations, we update the higher-level proxies $P_{l+1}$ based on the lower-level proxies $P_l$. Namely, we first update the cluster assignment $Q_l$ by assigning each lower-level proxy to its the nearest higher-level proxy $P_l^i$, i.e., $Q_l^i=\arg\min_j||P_l^i - P_{l+1}^j||_2^2$. Then, we update the higher-level proxies by taking the average of the lower-level proxies that are assigned to them, i.e., $P_{l+1}^j=\sum_{Q_l^i=j} P_l^i\big/|\mathbf{1}_{Q_l=j}|$, where $\mathbf{1}_{Q_l=j}$ is a vector with all ones at the location where $Q_l^i=j$ and zeros elsewhere. Note that our method can be easily adapted to work with other clustering algorithms such as Gaussian Mixture models \cite{yang2012robust}.
%In this way, our HPL drives the embeddings toward a more hierarchical structured space where not only small within-class distance, but also relatively small within-cluster distance.

\begin{table*}[t!]
\begin{center}
\begin{tabular}{l|ccc|ccc}
\toprule
\multirow{2}{*}{} & \multicolumn{3}{c|}{Concatenated (512-dim)} & \multicolumn{3}{c}{Separated (128-dim)} \\
\cline{2-4}\cline{5-7}
                  & MAP@R        & P@1          & RP           & MAP@R       & P@1         & RP          \\\hline
Contrastive \cite{hadsell2006dimensionality} & $44.51 \pm 0.28$ & $73.27 \pm 0.23$ & $47.45 \pm 0.28$ & $40.29 \pm 0.27$ & $69.28 \pm 0.22$ & $43.39 \pm 0.28$\\ 
% Triplet \cite{schroff2015facenet} & $43.72 \pm 0.21$ & $72.94 \pm 0.20$ & $46.79 \pm 0.21$ & $38.35 \pm 0.26$ & $68.03 \pm 0.26$ & $41.57 \pm 0.26$\\
CosFace \cite{wang2018cosface} & $46.92\pm 0.19$ & $75.79\pm 0.14$ & $49.77\pm 0.19$ & $40.69\pm 0.21$ & $70.71\pm 0.14$ & $43.56\pm 0.21$\\
ArcFace \cite{deng2019arcface} & $\mathbf{47.41\pm 0.40}$ & $\mathbf{76.20 \pm 0.27}$ & $\mathbf{50.27\pm 0.38}$ & $41.11\pm 1.22$ & $\mathbf{70.88\pm1.51}$ & $44.00\pm1.26$ \\
MS \cite{wang2019multi} & $46.42 \pm 1.67$ & $75.01 \pm 1.21$ & $49.45 \pm 1.67$ & $\mathbf{41.24 \pm 1.89}$ & $70.65 \pm 1.70$ & $\mathbf{44.40 \pm 1.85}$\\
SoftTriple \cite{qian2019softtriple} & $47.35\pm 0.19$ & $76.12\pm 0.17$ & $50.21\pm 0.18$ & $40.92\pm 0.20$ & $70.88\pm 0.20$ & $43.83\pm 0.20$\\
% Cont. + XBM \cite{wang2020cross} & $38.54$ & $68.78$ & $41.71$ & $30.41$ & $60.21$ & $33.47$\\
Proxy-NCA++ \cite{teh2020proxynca++} & $46.56$& $75.10$ & $49.50$ & $41.51$ & $70.43$ & $43.82$
\\
\hline
Proxy-NCA \cite{movshovitz2017no} & $47.22 \pm 0.21$& $75.89 \pm 0.17$ & $50.10 \pm 0.22$ & $41.74\pm 0.21$ & $71.30 \pm 0.20$ & $44.71 \pm 0.21$\\
HPL-NCA & $\mathbf{47.97}$ & $\mathbf{76.60}$ & $\mathbf{50.87}$ & $\mathbf{42.06}$ & $\mathbf{71.77}$ & $\mathbf{45.06}$\\
\hline 
Proxy Anchor \cite{kim2020proxy} & $47.88$ & $76.12$ & $50.82$ & $43.97$ & $72.79$ & $47.00$\\ 
HPL-PA & \textcolor{blue}{$\mathbf{49.07}$} & \textcolor{blue}{$\mathbf{76.97}$} & \textcolor{blue}{$\mathbf{51.97}$} & \textcolor{blue}{$\mathbf{45.11}$} & \textcolor{blue}{$\mathbf{73.84}$} & \textcolor{blue}{$\mathbf{48.10}$}\\
\bottomrule
\end{tabular}
\vskip 0.1in
\caption{\textbf{Results on the SOP dataset}. All hyper-parameters are selected by Bayesian Optimization as in \cite{musgrave2020metric}. Average performance across 10 runs and the 95\% confidence interval are reported whenever applicable. The best numbers in each block are marked as bold and the best numbers in the table are highlighted in blue. Cont. + XBM is not included because it failed to converge under this training setting.
}
\label{tb:RC_SOP}
\end{center}
\vspace{-0.2cm}
\end{table*}

\begin{table*}[t]
\begin{center}
\begin{tabular}{l|ccc|ccc}
\toprule
\multirow{2}{*}{} & \multicolumn{3}{c|}{Concatenated (512-dim)} & \multicolumn{3}{c}{Separated (128-dim)} \\
\cline{2-4}\cline{5-7}
                  & MAP@R        & P@1          & RP           & MAP@R       & P@1         & RP          \\\hline
Contrastive \cite{hadsell2006dimensionality} & $25.49 \pm 0.41$ &$81.57\pm 0.36$ & $35.72 \pm 0.35$ & $17.61 \pm 0.24$ & $69.44 \pm 0.24$ & $28.15 \pm 0.21$\\ 
% Triplet \cite{schroff2015facenet} & $22.13 \pm 0.45$ & $77.48 \pm 0.57$ & $32.85 \pm 0.45$ & $15.24 \pm 0.28$ & $63.87 \pm 0.41$ & $26.07 \pm 0.32$\\
CosFace \cite{wang2018cosface} & $26.86 \pm 0.22$ & $85.27 \pm 0.23$ & $36.72 \pm 0.20$ & $18.22 \pm 0.11$ & $\mathbf{74.13 \pm 0.21}$ & $28.49 \pm 0.14$\\
ArcFace \cite{deng2019arcface} & $27.22 \pm 0.30$ & $\mathbf{85.44 \pm 0.28}$& $37.02\pm 0.29$ & $17.11\pm 0.18$ & $72.10\pm 0.37$ & $27.29\pm 0.17$\\
MS \cite{wang2019multi} & $\mathbf{27.84 \pm 0.77}$ & $85.29 \pm 0.31$ & $\mathbf{37.96 \pm 0.63}$ & $\mathbf{18.77 \pm 0.69}$ & $73.73 \pm 0.96$ & $29.38 \pm 0.60$\\
SoftTriple \cite{qian2019softtriple} & $26.06 \pm 0.19$ & $83.66 \pm 0.22$ & $36.31 \pm 0.16$ & $18.72 \pm 0.11$ & $72.98 \pm 0.16$ & $\mathbf{29.39 \pm 0.10}$\\
Cont. + XBM \cite{wang2020cross} & $26.04 \pm 0.24$ & $83.67 \pm 0.35$ & $36.10 \pm 0.19$ & $18.07 \pm 0.11$ & $72.58 \pm 0.21$ & $28.55 \pm 0.10$\\
Proxy-NCA++ \cite{teh2020proxynca++} & $26.02 \pm 0.26$ & $82.09 \pm0.41$ & $36.31 \pm0.24$ & $18.63 \pm0.09$ & $70.60 \pm0.18$ & $29.35 \pm0.08$\\
\hline
Proxy-NCA \cite{movshovitz2017no} & $25.56 \pm 0.15$        & $83.20\pm 0.22$        & $35.80\pm 0.12$        & $18.32\pm 0.12$       & $\mathbf{73.34\pm 0.13}$       & $28.87\pm 0.11$       \\
HPL-NCA           & $\mathbf{27.47\pm 0.20}$        & $\mathbf{84.54\pm 0.25}$        & $\mathbf{37.56\pm 0.20}$        & $\mathbf{18.87\pm 0.13}$       & $72.27\pm 0.18$       & $\mathbf{29.45\pm 0.15}$\\
\hline
Proxy Anchor \cite{kim2020proxy} & $27.77 \pm 0.20$ & $86.38 \pm 0.15$ & $37.53 \pm 0.17$ & ${19.82 \pm 0.10}$ & \textcolor{blue}{$\mathbf{76.85 \pm 0.13}$} & $30.12 \pm 0.10$ \\
HPL-PA & \textcolor{blue}{$\mathbf{28.67 \pm0.22}$  }& \textcolor{blue}{$\mathbf{86.84 \pm0.31}$} & \textcolor{blue}{$\mathbf{38.36 \pm0.18}$} & \textcolor{blue}{$\mathbf{19.83 \pm 0.10}$} & $76.12 \pm 0.34$ & \textcolor{blue}{$\mathbf{30.13 \pm 0.09}$}\\

\bottomrule
\end{tabular}
\vskip 0.1in
\caption{\textbf{Results on the Cars-196 dataset}. All hyper-parameters are selected by Bayesian Optimization as in \cite{musgrave2020metric}. Average performance across 10 runs and the 95\% confidence interval are reported. %The best numbers in each block are marked as bold and the best numbers in the table are highlighted in blue.
}
% %%\vspace{-0.2cm}
\label{tb:RC_Cars}
\end{center}
\end{table*}
\begin{table*}[t!]
\begin{center}
\begin{tabular}{l|ccc|ccc}
\toprule
\multirow{2}{*}{} & \multicolumn{3}{c|}{Concatenated (512-dim)} & \multicolumn{3}{c}{Separated (128-dim)} \\
\cline{2-4}\cline{5-7}
                  & MAP@R        & P@1          & RP           & MAP@R       & P@1         & RP          \\\hline
Contrastive \cite{hadsell2006dimensionality} & $26.19 \pm 0.28$ &$67.21 \pm 0.49$ & $36.92 \pm 0.28$ & $20.73 \pm 0.19$ & $58.63 \pm 0.46$ & $31.48 \pm 0.19$\\ 
% Triplet \cite{schroff2015facenet} & $23.79 \pm 0.36$ & $64.40 \pm 0.38$ & $34.63 \pm 0.36$ & $18.80 \pm 0.27$ & $55.97 \pm 0.32$ & $29.60 \pm 0.28$\\
CosFace \cite{wang2018cosface} & $26.53 \pm 0.23$ & $67.19 \pm 0.37$ & $37.36 \pm 0.23$ & $21.25 \pm 0.18$ & $59.83 \pm 0.30$ & $32.07 \pm 0.19$\\
ArcFace \cite{deng2019arcface} & ${26.45\pm 0.20}$ & ${67.50 \pm 0.25}$ & ${37.31\pm 0.21}$ & $21.49\pm 0.16$ & ${60.17\pm0.32}$ & $32.37\pm0.17$ \\
MS \cite{wang2019multi} & $25.16 \pm 0.10$ & $65.93 \pm 0.16$ & $35.91 \pm 0.11$ & $20.58 \pm 0.09$ & $58.51 \pm 0.18$ & $31.36 \pm 0.10$\\
SoftTriple \cite{qian2019softtriple} & $25.64 \pm 0.21$ & $66.20 \pm 0.37$ & $36.46 \pm 0.20$ & $21.26 \pm 0.18$ & $59.55 \pm 0.35$ & $32.10 \pm 0.19$\\
Cont. + XBM \cite{wang2020cross} & \textcolor{blue}{$\mathbf{26.85 \pm 0.63}$} & \textcolor{blue}{$\mathbf{68.43 \pm 1.18}$} & \textcolor{blue}{$\mathbf{37.66 \pm 0.56}$} & $\mathbf{21.78 \pm 0.35}$ & $\mathbf{60.95 \pm 0.76}$ & $\mathbf{32.69 \pm 0.33}$\\
Proxy-NCA++ \cite{teh2020proxynca++} & $23.53\pm 0.12 $ & $64.69\pm 0.40$ & $34.37\pm 0.13$ & $18.76\pm 0.15$ & $57.13\pm 0.36$ & $29.52\pm 0.16$\\
\hline
Proxy-NCA \cite{movshovitz2017no} & $23.85 \pm0.24$        & $65.01 \pm0.27$        & $34.79 \pm0.26$        & $19.15 \pm0.15$       & $\mathbf{57.49 \pm0.35}$       & $29.99 \pm0.15$       \\
HPL-NCA & $\mathbf{24.95 \pm 0.21} $ & $\mathbf{65.22 \pm 0.23}$ & $\mathbf{35.70 \pm 0.21}$ & $\mathbf{20.04 \pm 0.21}$ & ${57.45 \pm 0.13} $ & $\mathbf{30.79 \pm 0.21}$ \\
\hline
Proxy Anchor \cite{kim2020proxy} & $26.47 \pm0.21$ & $67.64 \pm0.42$ & $37.29 \pm0.19$ & $21.57 \pm0.15$& $60.59 \pm0.24$ & $32.45 \pm0.15$\\ 
HPL-PA & $\mathbf{26.72\pm0.18}$ & $\mathbf{68.25\pm0.29}$ & $\mathbf{37.57\pm0.18}$& \textcolor{blue}{$\mathbf{21.90\pm0.19}$} & \textcolor{blue}{$\mathbf{61.31\pm0.25}$} & \textcolor{blue}{$\mathbf{32.81\pm0.19}$}\\
\bottomrule
\end{tabular}
\vskip 0.1in
\caption{\textbf{Results on the CUB dataset}. All hyper-parameters are selected by Bayesian Optimization as in \cite{musgrave2020metric}. Average performance across 10 runs and the 95\% confidence interval are reported. %The best numbers in each block are marked as bold and the best numbers in the table are highlighted in blue.
}
\label{tb:RC_CUB}
\end{center}
% %%\vspace{-.3cm}
\end{table*}

\section{Experiments}
To evaluate the proposed hierarchical proxy-based loss (HPL), we follow both a recently proposed standardized evaluation protocol proposed in \cite{musgrave2020metric} and the traditional evaluation protocol as in \cite{kim2020proxy,movshovitz2017no}. We compare our HPL against existing proxy-based losses including Proxy-NCA \cite{movshovitz2017no} and Proxy Anchor \cite{kim2020proxy} on several public benchmark datasets on image retrieval. We denote our HPL implemented based on Proxy-NCA loss and Proxy Anchor loss by HPL-NCA and HPL-PA, respectively. In addition, we also study the impact of different hyper-parameters to the performance of HPL as well as the effectiveness of the online clustering module of HPL.
Recall@$K$, Mean Average Precision at R (MAP@R) \cite{musgrave2020metric} and R-precision (RP) are used to measure the image retrieval performance

\subsection{Datasets}
Five popular benchmark datasets are used to evaluate our method: In-shop Clothes Retrieval (In-Shop) \cite{liu2016deepfashion}, Stanford Online Products (SOP) \cite{oh2016deep}, CUB-200-2011 (CUB) \cite{welinder2010caltech}, Cars-196 \cite{krause20133d} and iNaturalist \cite{Grant2018inat}.
\textbf{In-Shop} \cite{liu2016deepfashion} contains 72,712 clothes images of 7,986 classes, among which, the first 3,997 classes are used for training and the remaining 3985 classes for testing. Note that the testing data is split into a query set and a gallery set which contain 14,218 images and 12,612 images, respectively. 
\textbf{SOP} \cite{oh2016deep}
consists of 120,053 online product images of 22,634 classes and 24 super classes. We use the first 59,551 images (11,318 classes, 12 super classes) for training and the remaining 60,502 (11,316 classes, 12 super classes) for testing.
\textbf{Cars-196} \cite{krause20133d}
is composed of 196 car models (i.e., classes) with 16,185 images. We train the models on the first 98 classes (8,054 images) and test on the remaining 100 classes (8,131 images). 
\textbf{CUB-200-2011} \cite{welinder2010caltech}
contains 11,788 images of 200 bird species (i.e., classes). We train the models on the first 100 classes (5,864 images) and test on the remaining 100 classes (5,924 images).
\textbf{iNaturalist} \cite{Grant2018inat} is a fine-grained dataset of animal and plant species with human-curated hierarchy of categories. We use iNaturalist 2019\footnote{We choose iNaturalist 2019 over iNaturalist 2018 because the former is more balanced in its category hierarchy.} which contains 1,010 species, spanning 72 genera, combining a total of 268,243 images in the training and validation set. Each genus contains at least 10 species, making the dataset well-balanced in its category hierarchy. We follow \cite{brown2020smooth} and use the first 656 species (48 genera) for training and the remaining 354 species (24 genera) for testing. We will make the train/test splits publicly available.

\subsection{Implementation Details}
\myheading{Embedding network.}
We use the Inception network with batch normalization (BN-Inception) \cite{ioffe2015batch} and ResNet-50 \cite{he2016deep} as the backbone networks. Both backbone networks are pretrained on ImageNet \cite{deng2009imagenet} for the classification task. We append a max pooling layer in parallel with an average pooling layer to the penultimate layer of the backbone network as in \cite{kim2020proxy} and replace the final fully-connected (fc) layer of the backbone network with a new fc layer (i.e., embedding layer) which projects the network output to a desired embedding space.

\myheading{Structure of the hierarchical proxies.}
The structure of the hierarchical proxies varies for different datasets as different datasets have different hierarchical characteristics. However,
to keep the analysis simple, we evaluate our model in a simple conceptual structure, a two-level hierarchy. Namely, the number of fine (i.e, lower-level) proxies is set as the number of classes in the dataset. While the coarse (higher-level) proxies is chosen differently for different datasets as the number of classes in each dataset varies significantly. Please find the detailed study of the impact of the hierarchical structure at \Sref{sec:proxy_struct}.

\myheading{Training setting.}
Different from HTL \cite{ge2018deep} which initializes the hierarchical tree by clustering the whole datasets, which is expensive and infeasible for large datasets, we train the finest level of proxies $P_0$ and the embedding network with standard proxy losses (i.e., Proxy-NCA or Proxy Anchor) for the first 3 epochs to prevent the model from capturing wrong data hierarchy during the early stage of training.
The higher level of proxies are updated at every epoch. The loss weights in \Eref{eq:hpl} are set as $\omega_0=1.0$ and $\omega_1=0.1$.
Under the standardized evaluation protocol \cite{musgrave2020metric} all hyper-parameters of the models are determined using Bayesian Optimization \cite{snoek2012practical} and cross validation and the training terminates once the validation error plateaus. The embedding size is set to 128 and BN-Inception is used as the backbone networks. During testing, two modes are used for evaluation: \textit{Concatenated} where 128-dim embeddings of the 4 models trained with 4-fold cross validation are concatenated (512-dim) and \textit{Separated} where the 4 models are evaluated separately (128-dim) and the average performance is reported.
Under the traditional evaluation protocol, we train the baseline models using ResNet-50 as backbone with both Proxy-NCA loss and Proxy Anchor loss by following the standard hyper-parameters settings in \cite{movshovitz2017no} and \cite{kim2020proxy}, respectively. Each model is trained for 30 epochs with learning rate $10^{-4}$ and batch size 128. The embedding dimension is set to 512.
More details can be found in the supplement.

\subsection{Comparing Deep Metric Learning Models}
\begin{table}[t]
\begin{center}
\begin{tabular}{l|cc|cc}
\toprule
\multirow{2}{*}{} & \multicolumn{2}{c|}{In-Shop} & \multicolumn{2}{c}{SOP} \\
\cline{2-5}
                  & R@1 & R@10 & R@1 & R@10 \\\hline
Proxy-NCA & 87.21 & 96.57 & 77.63 & 89.29\\
HPL-NCA  & \textbf{88.70} & \textbf{96.83} & \textbf{80.13} & \textbf{91.07} \\
\hline 
Proxy Anchor & 89.85 & 97.14 & 79.38 & 90.44 \\
HPL-PA  & \textbf{92.46} & \textbf{97.97}  & \textbf{80.04} & \textbf{91.05} \\ 
\bottomrule
\end{tabular}
\vskip 0.1in
\caption{\textbf{Recall@$K$ (\%) on the In-Shop and SOP datasets}. ResNet-50 is used as the backbone and we set $|P_1|=500$ for both In-Shop and SOP.
}
\label{tb:inshop_sop}
\end{center}
\vspace{-0.2cm}
\end{table}

\myheading{Main Results.}
% Reality-Check evaluation protocol
To eliminate the bias incurred by the choice of hyper-parameters, we follow a stringent evaluation protocol proposed in \cite{musgrave2020metric} and compare our method based on Proxy Anchor loss (HPL-PA) with existing metric learning losses including Contrastive \cite{hadsell2006dimensionality}, %Triplet \cite{schroff2015facenet},
CosFace \cite{wang2018cosface}, ArcFace \cite{deng2019arcface}, MultiSimilarity (MS) \cite{wang2019multi}, SoftTriple \cite{qian2019softtriple}, Contrastive with Cross-Batch Memory (Cont. + XBM) \cite{wang2020cross} and Proxy-NCA++ \cite{teh2020proxynca++}.
\Tref{tb:RC_SOP}-\ref{tb:RC_CUB} shows the MAP@R, P@1 (i.e., Recall@1) and RP on the SOP, Cars-196 and CUB datasets, respectively. The results demonstrate the effectiveness of our proposed HPL. One can see that our HPL consistently improves both Proxy-NCA loss and Proxy Anchor loss in most metrics across all three datasets. Observe that  HPL improves the MAP@R of Proxy-NCA from 25.56\% to 27.47\% and from 47.22\% to 47.92\% on the Cars-196 and SOP datasets, respectively, making Proxy-NCA comparable with or even better than a recent proxy-based loss---Proxy Anchor. This suggests that the improvement of HPL over Proxy-NCA is significant.
Comparing with Proxy Anchor, our HPL-PA boosts the MAP@R by 1.19\%, 0.9\%, and 0.23\% on SOP, Cars-196 and CUB, respectively. This could imply that HPL tends to work better on large datasets with stronger data hierarchy (e.g., SOP) than on small ones (e.g., CUB and Cars-196) which lack hierarchy. Moreover, HPL-PA achieves  state-of-the-art performance on all three datasets comparing with all other approaches in most of the scenarios. In the CUB dataset, our HPL-PA is slightly worse than XBM with Contrastive loss, but HPL-PA is much more stable than XBM as one can see from the confidence interval.
% our HPL-NCA boosts the MAP@R by 1.91\%, P@1 by 1.34\% and RP by 1.76\% compared with Proxy-NCA; 
% HPL-PA boosts the MAP@R by 0.9\%, P@1 by 0.46\% and RP by 0.83\% compared with Proxy-PA.
%Although in the separated mode, the P@1 is slightly decreased, the MAP@R and RP which better reflect the overall quality of the learned embeddings are improved by 0.55\% and 0.58\%, respectively.
% Detailed configurations are discussed in the supplementary material.

We further compare our method with ResNet-50 as the backbone on two large datasets: In-Shop and SOP, using the traditional evaluation protocol. The number of coarse proxies in our HPL for the In-Shop dataset and SOP dataset are set to 500.
The results in \Tref{tb:inshop_sop} show that our HPL improves the traditional Proxy-NCA and Proxy Anchor on both benchmarks consistently. Especially, HPL-PA surpasses Proxy Anchor by 2.87\% on the In-Shop dataset and HPL-NCA outperforms Proxy-NCA by 2.50\% on the SOP dataset in Recall@1. This implies that the inclusion of class-shared information boosts the image retrieval performance and our HPL helps to capture this information as well as the class-discriminative features. More results can be found in the supplement.

\begin{figure}[t]
  \centering
  \includegraphics[width=1.\linewidth]{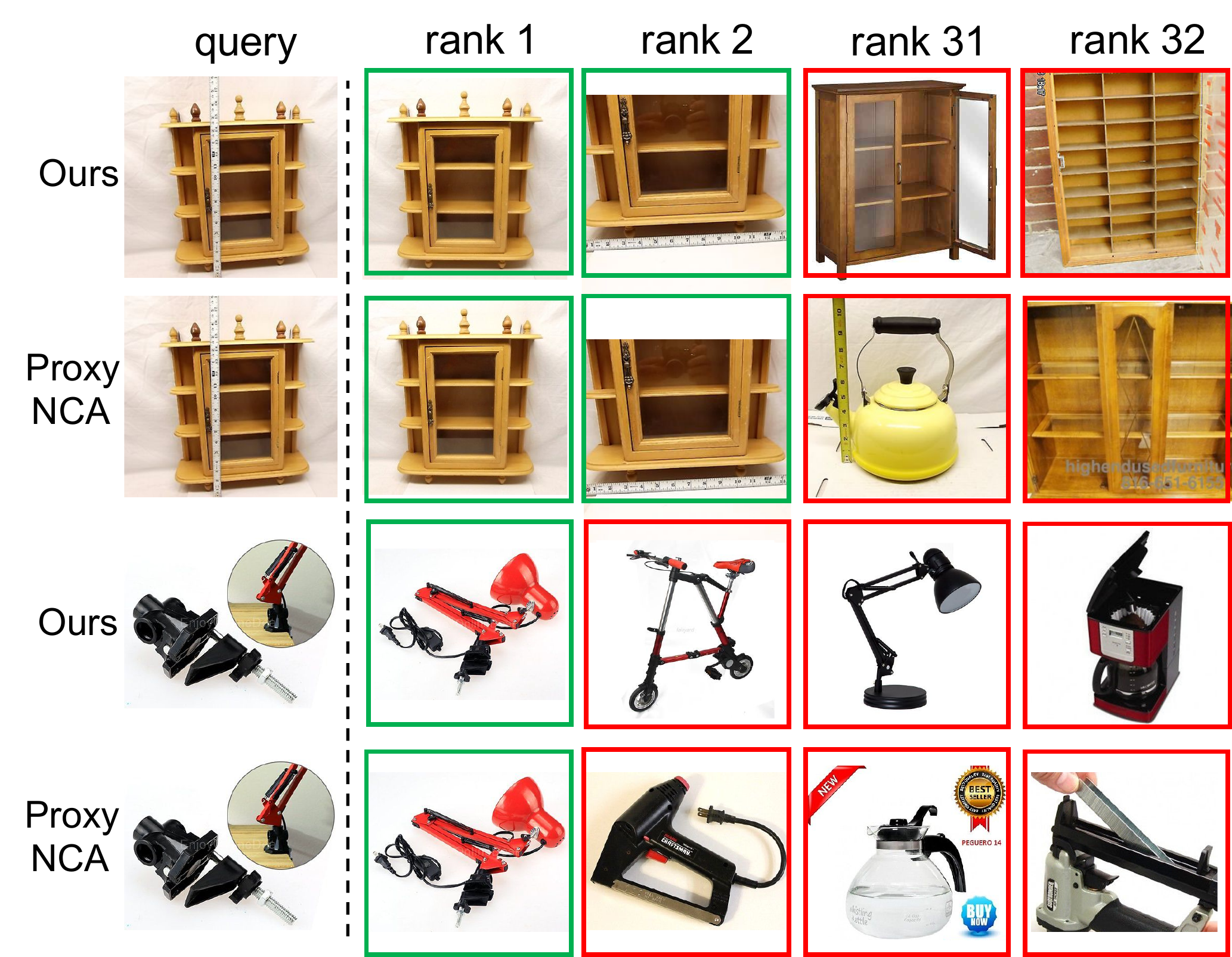}
  \caption{\textbf{Qualitative results on SOP}. Odd rows are the results of our HPL-NCA loss; even rows are the results from the Proxy-NCA loss. Green box indicates correct match, while red box indicates wrong match.}
  \label{fig:vis}
  \vspace{-.3cm}
\end{figure}

\myheading{Qualitative analysis.}
To further evaluate our method, we qualitatively compare our HPL-NCA with the Proxy-NCA loss by presenting the image retrieval results. In \Fref{fig:vis}, we present the rank-1, 2, 31, and 32 retrieval results on SOP to illustrate the improvement on the embeddings quality. One can see our HPL-NCA is better than Proxy-NCA in both overall quality of the image retrieval results and accuracy. Specifically, \Fref{fig:vis} reveals that despite both methods finding the correct matches, the retrieval results from our method are more similar to the queries. For example, on the left panel of \Fref{fig:vis} Proxy-NCA returns a yellow kettle as the 31th match, while our HPL-NCA returns a brown wood cabinet which is much more similar to the query image---a yellow wood hutch. Please find more qualitative results in the supplement.

% \subsection{Comparison to Existing Approaches}

\begin{figure}[t]
  \centering
  \includegraphics[width=1.\linewidth]{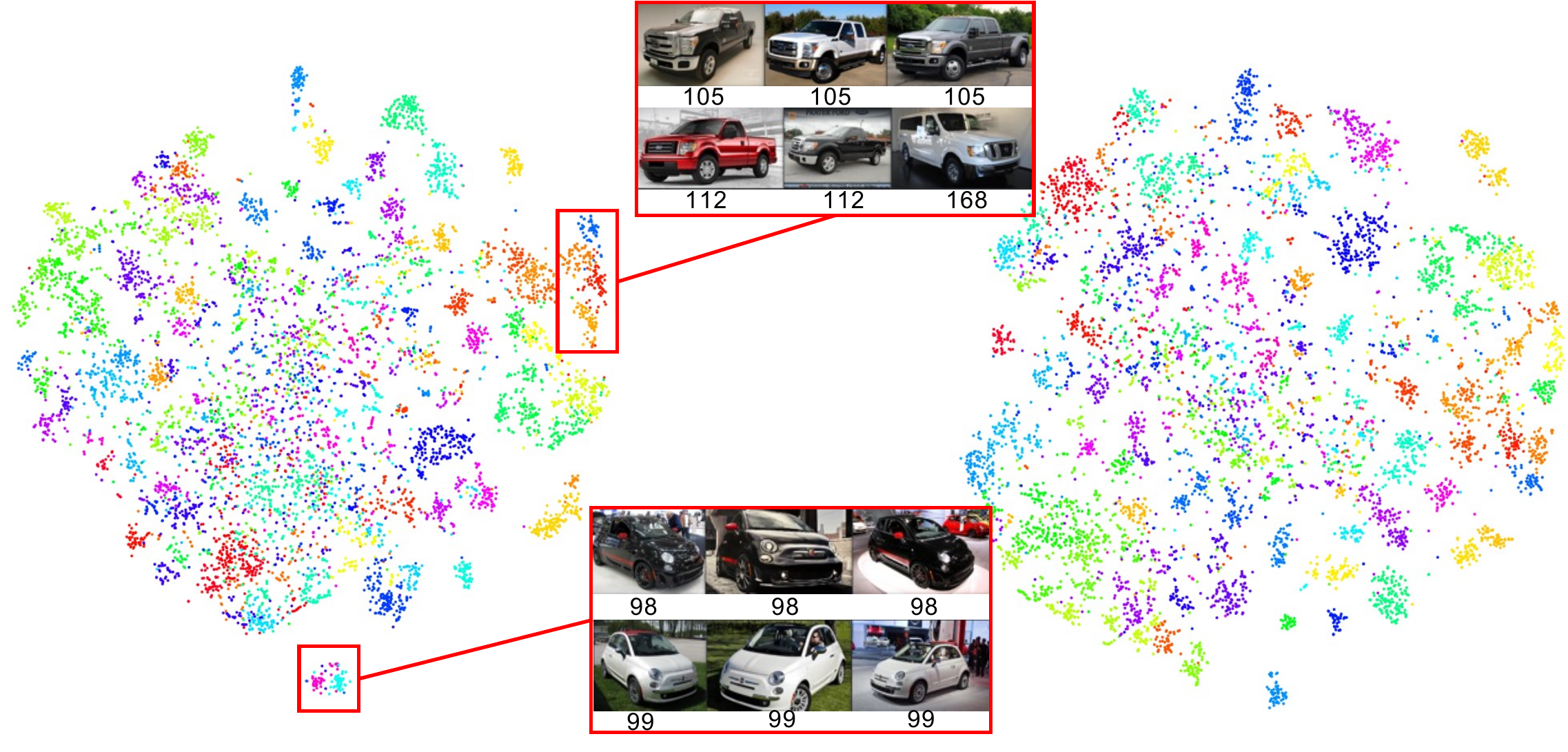}
  \caption{\textbf{Visualization of the embedding space.} We visualize the learned embeddings of the Cars-196 test set of our HPL-NCA (left) and Proxy-NCA (right) using t-SNE.}
  \label{fig:embdist}
  %\vspace{-.2cm}
\end{figure}

Furthermore, in Fig. \ref{fig:embdist} we visualize the embeddings of the test set of Cars-196 (98 classes) learned by our HPL-NCA (10 coarse proxies) and Proxy-NCA with t-SNE \cite{van2008visualizing}. As highlighted in red boxes, 
our method groups similar car categories (e.g., pickup trucks) into a larger cluster, while still maintaining good separability between classes. The learned common truck features help discern trucks from unseen car categories like SUVs. 

\subsection{Learning with a Human-curated Hierarchy}

Our method learns the data hierarchy in an unsupervised manner (see \Sref{sec:oc} for details). To demonstrate its effectiveness, we replace the online clustering module in Algorithm \ref{alg:oc} with a ground-truth class hierarchy. To this end, we use the SOP and the iNaturalist datasets where human-curated hierarchical labels are available. In particular, instead of performing online clustering with the fine proxies, we use a fixed assignment of the fine proxies to coarse proxies given by the ground-truth class hierarchy. The results in \Tref{tb:inat_sop} show that both HPL-NCA and HPL-NCA-GT (i.e., HPL-NCA with ground-truth hierarchy) outperform Proxy-NCA, and surprisingly, HPL-NCA surpasses HPL-NCA-GT even when given the same number of coarse proxies. This might be due to the fact that the human-curated category hierarchy may not fully reflect the visual similarity among classes, whereas, our method automatically learns the hierarchy based on visual similarity between classes, making it more favorable for metric learning. Please see the supplement for full results and further discussion.
% \begin{table}[t]
% \begin{center}
% \begin{tabular}{l|c|ccc}
% \toprule 
% Method & $|P_1|$ & R@1 & R@2 & R@4 \\ \midrule 
% Proxy-NCA   & -       & 51.32          & 62.56          & 72.53 \\
% HPL-NCA-GT  &48     & 51.63          & 63.00          & 72.77 \\
% HPL-NCA     &48    & 51.95 & 63.18 & 73.04\\
% HPL-NCA     &500    & \textbf{52.26} & \textbf{63.53} & \textbf{73.47}\\ %81.62 & 87.93 & 92.58
% \bottomrule
% \end{tabular}
% \vskip 0.1in
% \caption{Recall@$K$ (\%) on the iNaturalist dataset. HPL-NCA-GT denotes the HPL-NCA with ground-truth class hierarchy.
% }
% \label{tb:inat}
% \end{center}
% % %%\vspace{-0.2cm}
% \end{table}
% \begin{table}[t]
% \begin{center}
% \begin{tabular}{l|c|ccc}
% \toprule 
% Method & $|P_1|$ & R@1 & R@10 & R@100 \\ \midrule 
% Proxy-NCA   & - & $77.63$ & $89.29$ & $94.58$ \\
% HPL-NCA-GT  &12      & $78.69$ & $90.44$ & $95.72$ \\
% HPL-NCA     &12    & $79.33$& $90.71$& $95.63$\\
% HPL-NCA     &100    & \textbf{79.83} & \textbf{91.06} & \textbf{96.07}\\ %81.62 & 87.93 & 92.58
% \bottomrule
% \end{tabular}
% \vskip 0.1in
% \caption{Recall@$K$ (\%) on the SOP dataset. HPL-NCA-GT denotes the HPL-NCA with ground-truth class hierarchy.
% }
% \label{tb:sop_gt}
% \end{center}
% % %%\vspace{-0.2cm}
% \end{table}

\begin{figure}[t]
  \centering
  \includegraphics[width=1.0\linewidth]{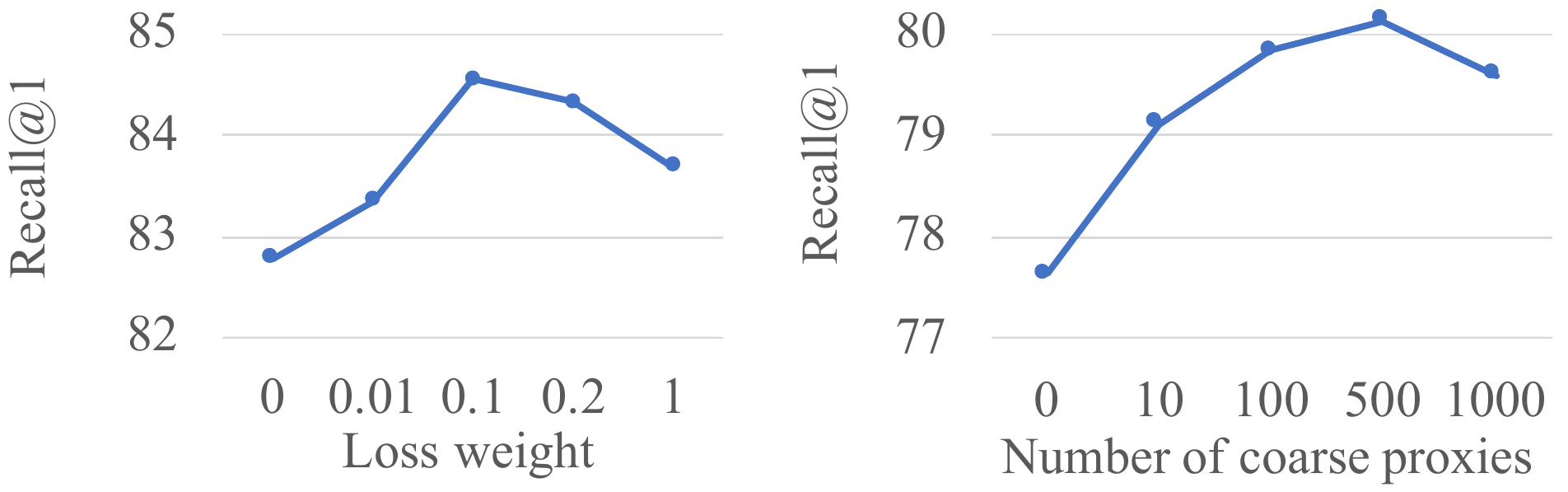}
  \caption{\textbf{Impact of loss weight (left) and number of coarse proxies (right)}. % We train the models with HPL-NCA loss and two proxy levels. 
  Left: Recall@1 on the Cars-196 dataset of models that are trained with 20 coarse proxies but different loss weights $\omega_1$. Right: Recall@1 on the SOP dataset of models that are trained with different number of coarse proxies ($\omega_1=0.1$).}
  \label{fig:hp_cars}
  \vspace{-0.2cm}
\end{figure}

\subsection{Impact of Hyperparameters} \label{sec:proxy_struct}

\myheading{Loss weight.}
In \Eref{eq:hpl}, we combine losses contributed by different levels of proxies using a weighted summation with the weights $\omega$ as hyper-parameters. We fix $\omega_0=1.0$ and train the embedding networks on Cars-196 by varying $\omega_1$. As shown in \Fref{fig:hp_cars}, the model performs best when $\omega_1=0.1$. Note that when $\omega_1=0$ our HPL-NCA loss is equivalent to the standard Proxy-NCA loss. Hence, for all $\omega_1>0$, our HPL-NCA loss consistently beats the standard Proxy-NCA loss (i.e., $\omega_1=0$). This further demonstrates the superiority of our HPL over standard proxy-based losses.

\myheading{Hierarchical structure.}
To better study the impact of the hierarchical structure, we use the SOP dataset which contains 11,316 training classes---much more than the other four datasets. Specifically, we train models on the SOP dataset using HPL-NCA loss with a variable number of coarse proxies $|P_1|=$ 0, 10, 100, 500 and 1000. Note that when $|P_1|=0$ HPL-NCA is equivalent to Proxy-NCA. The results in \Fref{fig:hp_cars} show that
our method outperforms the baseline under different numbers of coarse proxies and the performance of our method is shown to be robust with respect to the number of coarse proxies. In general, having more coarse proxies is beneficial as more accurate class-shared information can be learned from the coarse proxies. However, an excessively large number of coarse proxies would possibly reduce the strength of the signal shared across classes. 
More hyperparameter analysis can be found in the supplement.

% \myheading{Impact of coarse proxies updating interval.}

\begin{table}[t]
\begin{center}
\begin{tabular}{l|c|c|c|c}
\toprule
\multirow{2}{*}{} & \multicolumn{2}{c|}{SOP} & \multicolumn{2}{c}{iNaturalist} \\
\cline{2-5}
                  & $|P_1|$ & R@1 & $|P_1|$ & R@1 \\\hline
Proxy-NCA   & - & $77.63$ & - & $51.32$ \\
HPL-NCA-GT  & $12$  & $78.69$ & $48$ & $51.63$ \\
HPL-NCA     &$12$  & $79.33$& $48$& $51.95$\\
HPL-NCA     &$500$    & $\mathbf{80.13}$ & 500 & $\mathbf{52.26}$\\
\bottomrule
\end{tabular}
\vskip 0.1in
\caption{\textbf{Learned hierarchy vs human-curated hierarchy}. HPL-NCA-GT denotes PL-NCA with ground-truth class hierarchy.
}
\label{tb:inat_sop}
\end{center}
\vspace{-0.2cm}
\end{table}
\section{Conclusion}
In this paper, we have demonstrated the effectiveness of a hierarchical proxy-based loss (HPL), which enforces a hierarchical structure on the learnable proxies. In this way, we are able to not only learn class-discriminative information, but also capture the features that are shared across classes, which improves the generalizability of the learned embeddings. As a result, our proposed HPL improves standard proxy-based losses, especially on large datasets where a clear hierarchical structure exists in the data space. In future work, we will explore more configurations of our method, e.g., instead of using a simple two-level hierarchy and k-means where the number of clusters has to be predefined, we can use hierarchical clustering algorithms to automatically learn a multi-level hierarchy to better fits the data.

\myheading{Acknowledgements}. This project is supported by US National Science Foundation Award IIS-1763981, the Partner University Fund, the SUNY2020 Infrastructure Transportation Security Center, and a gift from Adobe.

{\small
\bibliographystyle{ieee_fullname}
\bibliography{egbib}

\begin{thebibliography}{10}\itemsep=-1pt

\bibitem{aziere2019ensemble}
Nicolas Aziere and Sinisa Todorovic.
\newblock Ensemble deep manifold similarity learning using hard proxies.
\newblock In {\em Proceedings of the IEEE Conference on Computer Vision and
  Pattern Recognition}, pages 7299--7307, 2019.

\bibitem{boudiaf2020unifying}
Malik Boudiaf, J{\'e}r{\^o}me Rony, Imtiaz~Masud Ziko, Eric Granger, Marco
  Pedersoli, Pablo Piantanida, and Ismail~Ben Ayed.
\newblock A unifying mutual information view of metric learning: cross-entropy
  vs. pairwise losses.
\newblock In {\em European Conference on Computer Vision}, pages 548--564.
  Springer, 2020.

\bibitem{brown2020smooth}
Andrew Brown, Weidi Xie, Vicky Kalogeiton, and Andrew Zisserman.
\newblock Smooth-ap: Smoothing the path towards large-scale image retrieval.
\newblock In {\em European Conference on Computer Vision}, pages 677--694.
  Springer, 2020.

\bibitem{chen2017beyond}
Weihua Chen, Xiaotang Chen, Jianguo Zhang, and Kaiqi Huang.
\newblock Beyond triplet loss: a deep quadruplet network for person
  re-identification.
\newblock In {\em Proceedings of the IEEE Conference on Computer Vision and
  Pattern Recognition}, pages 403--412, 2017.

\bibitem{dai2019batch}
Zuozhuo Dai, Mingqiang Chen, Xiaodong Gu, Siyu Zhu, and Ping Tan.
\newblock Batch dropblock network for person re-identification and beyond.
\newblock In {\em Proceedings of the IEEE International Conference on Computer
  Vision}, pages 3691--3701, 2019.

\bibitem{deng2009imagenet}
Jia Deng, Wei Dong, Richard Socher, Li-Jia Li, Kai Li, and Li Fei-Fei.
\newblock Imagenet: A large-scale hierarchical image database.
\newblock In {\em 2009 IEEE Conference on Computer Vision and Pattern
  Recognition}, pages 248--255. Ieee, 2009.

\bibitem{deng2019arcface}
Jiankang Deng, Jia Guo, Niannan Xue, and Stefanos Zafeiriou.
\newblock Arcface: Additive angular margin loss for deep face recognition.
\newblock In {\em Proceedings of the IEEE/CVF Conference on Computer Vision and
  Pattern Recognition}, pages 4690--4699, 2019.

\bibitem{elezi2020group}
Ismail Elezi, Sebastiano Vascon, Alessandro Torcinovich, Marcello Pelillo, and
  Laura Leal-Taix{\'e}.
\newblock The group loss for deep metric learning.
\newblock In {\em European Conference on Computer Vision}, pages 277--294.
  Springer, 2020.

\bibitem{ge2018deep}
Weifeng Ge, Weilin Huang, Dengke Dong, and Matthew~R Scott.
\newblock Deep metric learning with hierarchical triplet loss.
\newblock In {\em Proceedings of the European Conference on Computer Vision},
  pages 269--285, 2018.

\bibitem{ge2020self}
Yixiao Ge, Haibo Wang, Feng Zhu, Rui Zhao, and Hongsheng Li.
\newblock Self-supervising fine-grained region similarities for large-scale
  image localization.
\newblock In {\em European Conference on Computer Vision}, 2020.

\bibitem{hadsell2006dimensionality}
Raia Hadsell, Sumit Chopra, and Yann LeCun.
\newblock Dimensionality reduction by learning an invariant mapping.
\newblock In {\em Proceedings of the IEEE Conference on Computer Vision and
  Pattern Recognition}, volume~2, pages 1735--1742. IEEE, 2006.

\bibitem{he2016deep}
Kaiming He, Xiangyu Zhang, Shaoqing Ren, and Jian Sun.
\newblock Deep residual learning for image recognition.
\newblock In {\em Proceedings of the IEEE Conference on Computer Vision and
  Pattern Recognition}, pages 770--778, 2016.

\bibitem{ho2019pies}
Chih-Hui Ho, Pedro Morgado, Amir Persekian, and Nuno Vasconcelos.
\newblock Pies: Pose invariant embeddings.
\newblock In {\em Proceedings of the IEEE/CVF Conference on Computer Vision and
  Pattern Recognition}, pages 12377--12386, 2019.

\bibitem{Grant2018inat}
Grant~Van Horn, Oisin~Mac Aodha, Yang Song, Yin Cui, Chen Sun, Alexander
  Shepard, Hartwig Adam, Pietro Perona, and Serge~J. Belongie.
\newblock The inaturalist species classification and detection dataset.
\newblock In {\em Proceedings of the IEEE Conference on Computer Vision and
  Pattern Recognition}, pages 8769--8778, 2018.

\bibitem{ioffe2015batch}
Sergey Ioffe and Christian Szegedy.
\newblock Batch normalization: Accelerating deep network training by reducing
  internal covariate shift.
\newblock In {\em International conference on machine learning}, pages
  448--456. PMLR, 2015.

\bibitem{kim2018hierarchy}
Hyo~Jin Kim and Jan-Michael Frahm.
\newblock Hierarchy of alternating specialists for scene recognition.
\newblock In {\em Proceedings of the European Conference on Computer Vision
  (ECCV)}, pages 451--467, 2018.

\bibitem{kim2020proxy}
Sungyeon Kim, Dongwon Kim, Minsu Cho, and Suha Kwak.
\newblock Proxy anchor loss for deep metric learning.
\newblock In {\em Proceedings of the IEEE/CVF Conference on Computer Vision and
  Pattern Recognition}, pages 3238--3247, 2020.

\bibitem{krause20133d}
Jonathan Krause, Michael Stark, Jia Deng, and Li Fei-Fei.
\newblock 3d object representations for fine-grained categorization.
\newblock In {\em Proceedings of the IEEE international conference on computer
  vision workshops}, pages 554--561, 2013.

\bibitem{levi2020reducing}
Elad Levi, Tete Xiao, Xiaolong Wang, and Trevor Darrell.
\newblock Reducing class collapse in metric learning with easy positive
  sampling.
\newblock {\em arXiv preprint arXiv:2006.05162}, 2020.

\bibitem{liu2016deepfashion}
Ziwei Liu, Ping Luo, Shi Qiu, Xiaogang Wang, and Xiaoou Tang.
\newblock Deepfashion: Powering robust clothes recognition and retrieval with
  rich annotations.
\newblock In {\em Proceedings of the IEEE Conference on Computer Vision and
  Pattern Recognition}, pages 1096--1104, 2016.

\bibitem{milbich2020diva}
Timo Milbich, Karsten Roth, Homanga Bharadhwaj, Samarth Sinha, Yoshua Bengio,
  Bj{\"o}rn Ommer, and Joseph~Paul Cohen.
\newblock Diva: Diverse visual feature aggregation for deep metric learning.
\newblock In {\em European Conference on Computer Vision}, pages 590--607.
  Springer, 2020.

\bibitem{movshovitz2017no}
Yair Movshovitz-Attias, Alexander Toshev, Thomas~K Leung, Sergey Ioffe, and
  Saurabh Singh.
\newblock No fuss distance metric learning using proxies.
\newblock In {\em Proceedings of the IEEE International Conference on Computer
  Vision}, pages 360--368, 2017.

\bibitem{murthy2016deep}
Venkatesh~N Murthy, Vivek Singh, Terrence Chen, R Manmatha, and Dorin
  Comaniciu.
\newblock Deep decision network for multi-class image classification.
\newblock In {\em Proceedings of the IEEE conference on computer vision and
  pattern recognition}, pages 2240--2248, 2016.

\bibitem{musgrave2020metric}
Kevin Musgrave, Serge Belongie, and Ser-Nam Lim.
\newblock A metric learning reality check.
\newblock In {\em European Conference on Computer Vision}, pages 681--699.
  Springer, 2020.

\bibitem{oh2016deep}
Hyun Oh~Song, Yu Xiang, Stefanie Jegelka, and Silvio Savarese.
\newblock Deep metric learning via lifted structured feature embedding.
\newblock In {\em Proceedings of the IEEE Conference on Computer Vision and
  Pattern Recognition}, pages 4004--4012, 2016.

\bibitem{qian2019softtriple}
Qi Qian, Lei Shang, Baigui Sun, Juhua Hu, Hao Li, and Rong Jin.
\newblock Softtriple loss: Deep metric learning without triplet sampling.
\newblock In {\em Proceedings of the IEEE International Conference on Computer
  Vision}, pages 6450--6458, 2019.

\bibitem{roth2019mic}
Karsten Roth, Biagio Brattoli, and Bjorn Ommer.
\newblock Mic: Mining interclass characteristics for improved metric learning.
\newblock In {\em Proceedings of the IEEE International Conference on Computer
  Vision}, pages 8000--8009, 2019.

\bibitem{roweis2004neighbourhood}
Sam Roweis, Geoffrey Hinton, and Ruslan Salakhutdinov.
\newblock Neighbourhood component analysis.
\newblock {\em Advances in Neural Information Processing Systems}, 17:513--520,
  2004.

\bibitem{rumelhart1985learning}
David~E Rumelhart, Geoffrey~E Hinton, and Ronald~J Williams.
\newblock Learning internal representations by error propagation.
\newblock Technical report, California Univ San Diego La Jolla Inst for
  Cognitive Science, 1985.

\bibitem{ILSVRC15}
Olga Russakovsky, Jia Deng, Hao Su, Jonathan Krause, Sanjeev Satheesh, Sean Ma,
  Zhiheng Huang, Andrej Karpathy, Aditya Khosla, Michael Bernstein,
  Alexander~C. Berg, and Li Fei-Fei.
\newblock {ImageNet Large Scale Visual Recognition Challenge}.
\newblock {\em International Journal of Computer Vision (IJCV)},
  115(3):211--252, 2015.

\bibitem{sanakoyeu2019divide}
Artsiom Sanakoyeu, Vadim Tschernezki, Uta Buchler, and Bjorn Ommer.
\newblock Divide and conquer the embedding space for metric learning.
\newblock In {\em Proceedings of the IEEE/CVF Conference on Computer Vision and
  Pattern Recognition}, pages 471--480, 2019.

\bibitem{schroff2015facenet}
Florian Schroff, Dmitry Kalenichenko, and James Philbin.
\newblock Facenet: A unified embedding for face recognition and clustering.
\newblock In {\em Proceedings of the IEEE Conference on Computer Vision and
  Pattern Recognition}, pages 815--823, 2015.

\bibitem{snoek2012practical}
Jasper Snoek, Hugo Larochelle, and Ryan~Prescott Adams.
\newblock Practical bayesian optimization of machine learning algorithms.
\newblock {\em Advances in Neural Information Processing Systems}, 2012.

\bibitem{teh2020proxynca++}
Eu~Wern Teh, Terrance DeVries, and Graham~W Taylor.
\newblock Proxynca++: Revisiting and revitalizing proxy neighborhood component
  analysis.
\newblock In {\em European Conference on Computer Vision}. Springer, 2020.

\bibitem{van2008visualizing}
Laurens Van~der Maaten and Geoffrey Hinton.
\newblock Visualizing data using t-sne.
\newblock {\em Journal of machine learning research}, 9(11), 2008.

\bibitem{waltner2019hibster}
Georg Waltner, Michael Opitz, Horst Possegger, and Horst Bischof.
\newblock Hibster: Hierarchical boosted deep metric learning for image
  retrieval.
\newblock In {\em 2019 IEEE Winter Conference on Applications of Computer
  Vision (WACV)}, pages 599--608. IEEE, 2019.

\bibitem{wang2018cosface}
Hao Wang, Yitong Wang, Zheng Zhou, Xing Ji, Dihong Gong, Jingchao Zhou, Zhifeng
  Li, and Wei Liu.
\newblock Cosface: Large margin cosine loss for deep face recognition.
\newblock In {\em Proceedings of the IEEE Conference on Computer Vision and
  Pattern Recognition}, pages 5265--5274, 2018.

\bibitem{wang2019multi}
Xun Wang, Xintong Han, Weilin Huang, Dengke Dong, and Matthew~R Scott.
\newblock Multi-similarity loss with general pair weighting for deep metric
  learning.
\newblock In {\em Proceedings of the IEEE Conference on Computer Vision and
  Pattern Recognition}, pages 5022--5030, 2019.

\bibitem{wang2020cross}
Xun Wang, Haozhi Zhang, Weilin Huang, and Matthew~R Scott.
\newblock Cross-batch memory for embedding learning.
\newblock In {\em Proceedings of the IEEE/CVF Conference on Computer Vision and
  Pattern Recognition}, pages 6388--6397, 2020.

\bibitem{welinder2010caltech}
Peter Welinder, Steve Branson, Takeshi Mita, Catherine Wah, Florian Schroff,
  Serge Belongie, and Pietro Perona.
\newblock Caltech-ucsd birds 200.
\newblock 2010.

\bibitem{wu2017sampling}
Chao-Yuan Wu, R Manmatha, Alexander~J Smola, and Philipp Krahenbuhl.
\newblock Sampling matters in deep embedding learning.
\newblock In {\em Proceedings of the IEEE International Conference on Computer
  Vision}, pages 2840--2848, 2017.

\bibitem{yan2015hd}
Zhicheng Yan, Hao Zhang, Robinson Piramuthu, Vignesh Jagadeesh, Dennis DeCoste,
  Wei Di, and Yizhou Yu.
\newblock Hd-cnn: hierarchical deep convolutional neural networks for large
  scale visual recognition.
\newblock In {\em Proceedings of the IEEE international conference on computer
  vision}, pages 2740--2748, 2015.

\bibitem{yang2012robust}
Miin-Shen Yang, Chien-Yo Lai, and Chih-Ying Lin.
\newblock A robust em clustering algorithm for gaussian mixture models.
\newblock {\em Pattern Recognition}, 45(11):3950--3961, 2012.

\bibitem{zhan2020online}
Xiaohang Zhan, Jiahao Xie, Ziwei Liu, Yew-Soon Ong, and Chen~Change Loy.
\newblock Online deep clustering for unsupervised representation learning.
\newblock In {\em Proceedings of the IEEE/CVF Conference on Computer Vision and
  Pattern Recognition}, pages 6688--6697, 2020.

\end{thebibliography}
}

\end{document}